\documentclass[letterpaper,twocolumn,10pt]{article}
\usepackage{usenix-2020-09}

\usepackage{packages/packages}
\usepackage{packages/macros}

\usepackage{filecontents}

\begin{document}

\date{}
\title{\ours: A Robotics Foundation Model Serving System for Robot Factories}

\author{
\begin{tabular}{c}
Wenqi Jiang\textsuperscript{1}
\quad
Jason Clemons\textsuperscript{1}
\quad
Rowland O'Flaherty\textsuperscript{1}
\quad
Hugo Hadfield\textsuperscript{1}
\\
Alperen Degirmenci\textsuperscript{1}
\quad
Shuran Song\textsuperscript{1,2}
\quad
Yashraj Narang\textsuperscript{1}
\quad
Christos Kozyrakis\textsuperscript{1,2}
\\[0.6em]
\textsuperscript{1}NVIDIA Research
\quad
\textsuperscript{2}Stanford University
\end{tabular}
}

\maketitle

\begin{abstract}
    Robotics foundation models (RFMs) are making general-purpose robots increasingly practical for factory deployments.
While RFM serving systems are central to this vision, existing systems are largely shaped by a single-robot, single-model assumption: inference is treated as an edge-computing problem handled by an on-robot or dedicated nearby GPU, and the serving objective is to minimizing the latency of a single action model.
In this paper, we propose \ours, the \underline{R}obotics \underline{O}riented \underline{S}erving \underline{A}rchitecture, an RFM serving system for robot factories designed around three key principles.
First, \ours adopts \textit{shared GPU-pool serving}, allowing a fleet of robots to access powerful server-class GPUs over the network in order to improve inference performance, battery duration, and GPU utilization.
Second, \ours provides a \textit{robotics-aware programming abstraction and system design} that supports multi-model pipelines, per-task performance requirements, and failure handling.
Third, \ours uses \textit{factory-objective-driven scheduling} to maximize SLO-qualified factory productivity rather than minimizing individual request latency.
We implement \ours on top of Ray Serve for distributed orchestration, with vLLM, PyTorch, and JAX as model-serving backends, and evaluate it on both real robots and synthetic large-scale workloads.
The results show that \ours improves factory productivity by up to 12.06$\times$ over conventional dedicated serving systems.

\end{abstract}

\section{Introduction}
\label{sec:intro}

Embodied AI is emerging as a central frontier in the next phase of AI, promising physical agents that can perceive complex environments, reason over long-horizon tasks, and execute precise actions in the real world.
Central to this shift are robotics foundation models (RFMs), including Vision-Language-Action (VLA) models~\cite{zhao2023learning, black2024pi0, kim2024openvla, zitkovich2023rt, bjorck2025gr00t}, World Action Models (WAMs)~\cite{ye2026world, agarwal2026cosmos, zhang2026qwen}, and the surrounding reasoning and control models that support robotic execution.
By integrating semantic language understanding with sensor inputs in the action-generation loop, these models have demonstrated unprecedented generalization capabilities across diverse tasks, from manipulation to navigation~\cite{intelligence2504pi0, amin2025pi, huang2026fast, team2025gemini, bjorck2025gr00t, shah2023lm, wang2025alpamayo}.

\begin{figure}[t]
  \centering
  \includegraphics[width=\linewidth]{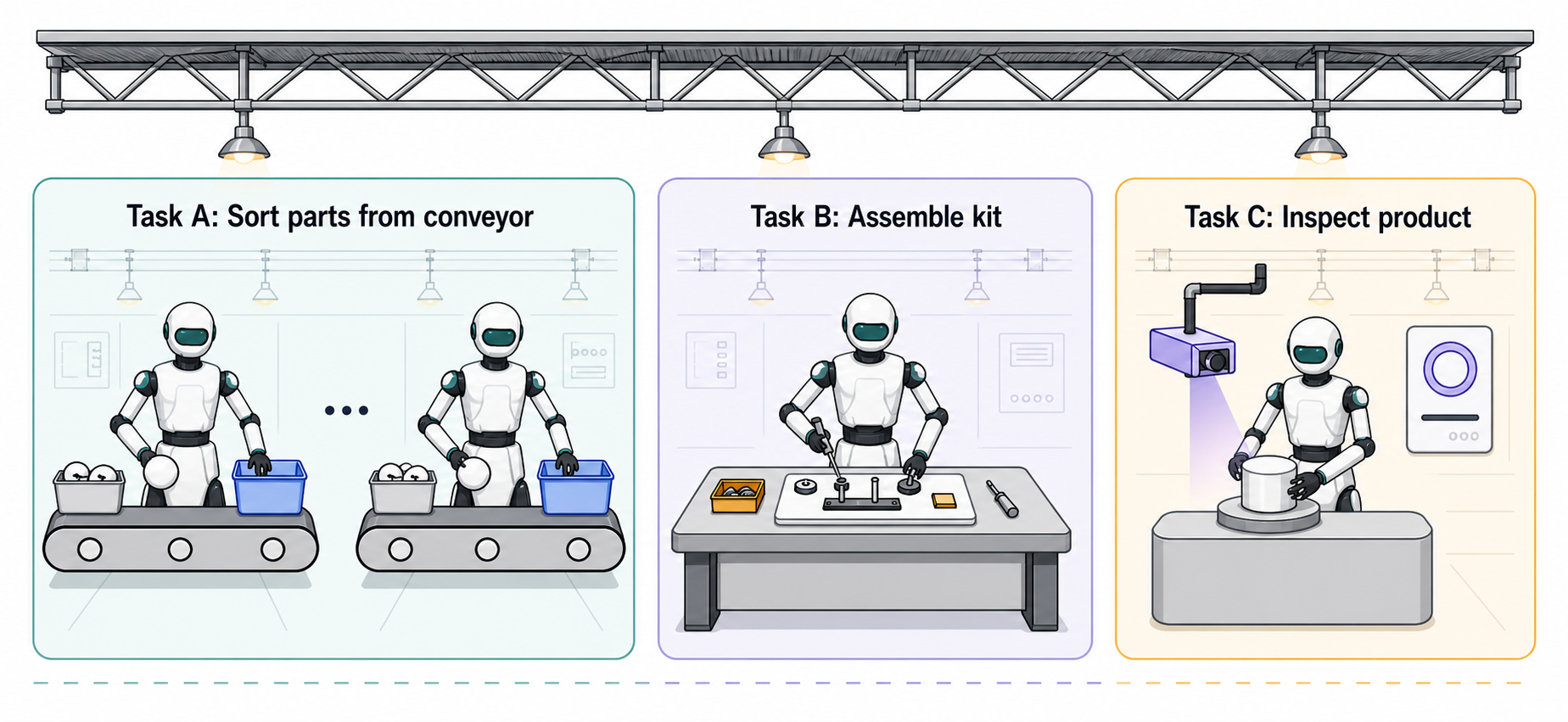}
  \vspace{-2em}
  \caption{Robots working on various tasks in a factory.}
  \vspace{-1em}
  \label{fig:factory}
\end{figure}

A promising use case for RFMs is the deployment of general-purpose robots, such as humanoids and manipulators, in highly-automated \textit{robot factories}.
For example, Figure AI has explored humanoid deployments in BMW manufacturing facilities for automotive assembly-line tasks~\cite{figure2025bmwdeployment}, Tesla is pursuing a similar vision with Optimus~\cite{vaziri2026teslaoptimusfremont}, and Amazon continues to scale warehouse robot fleets for inventory movement, sortation, and package handling~\cite{park2025vulcan, agaskar2025deepfleet}.

RFM serving systems are central to this vision, since such deployments rely on continuous streams of model inference to generate robot actions and monitor task execution.
Although recent work has begun to optimize RFM inference from both algorithmic and systems perspectives\cite{shukor2025smolvla, bjorck2025gr00t, dai2026kairos, ma2025running, ye2026world}, we argue that the RFM serving problem for robot factories is neither well defined nor well solved, due in part to two pervasive misconceptions:

\niparagraph{Misconception 1: robotics foundation model serving is strictly an edge computing problem.}
The prevailing deployment paradigm couples each robot with an onboard System-on-Chip (SoC) or a dedicated GPU server.
However, one SoC per robot is not only expensive but also provides limited performance because robot platforms often have tight power and thermal budgets~\cite{huang2026roboticrevolution}.
Offloading inference to a dedicated GPU server leads to better performance~\cite{jiang2026fast}, but this one-to-one serving paradigm still under-utilizes GPUs: a GPU serving only one robot cannot exploit inter-robot request batching, and in the common synchronous inference setting~\cite{black2024pi0,bjorck2025gr00t}, the GPU remains idle while waiting for the robot to finish executing the current action and issue the next inference request.

\niparagraph{Misconception 2: the main serving objective is to minimize inference latency for a single action model.} 
First, reliable robotic task execution often requires additional models beyond action prediction, such as high-level planning, safety checking, and task progression monitoring models, which should also be considered in the serving system design.
Second, action model inference latency should not be the sole optimization objective.
For example, a robot factory may aim to maximize productivity, measured as total action throughput across robots, while satisfying the latency service-level objectives (SLOs) of all required model components.

\niparagraph{\ours: efficient, reliable, and model-agnostic RFM serving.}
In this paper, we rethink the robot-factory serving problem through three questions: (a) what requirements arise from the robotics perspective, (b) how should serving systems be designed to satisfy these requirements, and (c) what optimization objectives should guide scheduling decisions?
We argue that future RFM serving systems for robot factories should be built around three core principles, regardless of the specific models they serve:

\niparagraph{Principle 1: server-scale, shared RFM serving infrastructure.}
While conventional deployments equip each robot with an onboard GPU  or a dedicated edge server, we argue that future robotic factories will increasingly rely on {shared GPU-pool serving systems}, where multiple robots share a centralized pool of compute resources.
This transition is motivated by three key advantages (A1--A3).

\begin{itemize}[wide=0pt, topsep=4pt, itemsep=4pt, parsep=0pt, partopsep=0pt] %

\item[(A1)] \textit{Better inference performance through access to data-center-class hardware.}
On-robot accelerators are significantly less powerful than modern data-center GPUs. 
For example, NVIDIA's Jetson Thor and B100 are manufactured in the same generation of technology, yet the latter provided 29$\times$ memory bandwidth and 3.5$\times$ compute capability (fp8) compared to the former.
As robotics foundation models continue to grow in size, the latency reduction achieved through faster inference can outweigh the additional network communication delay, particularly in stable industrial networks~\cite{jiang2026fast}.
Instead, on-robot SoCs can be used for low-compute, high-frequency actuator control and safety fallback operations.

\item[(A2)]  \textit{Extended operational duration for battery-powered robots.}
Many mobile robots rely on battery-powered platforms~\cite{huang2026roboticrevolution}.
Hosting compute-intensive inference locally increases power consumption and shortens operational duration. 
For example, Figure AI's Figure 02 robot features a 2.25 kWh battery yielding a runtime of roughly five hours, with its dual onboard NVIDIA RTX GPUs comprising up to half of the system's total power.
Offloading inference to centralized infrastructure reduces onboard energy usage, thereby extending battery duration and improving robot availability without requiring frequent recharging cycles.

\item[(A3)] \textit{Improved hardware utilization and cost efficiency.}
With requests arriving from multiple robots, a shared serving system enables inter-robot batching, thereby improving GPU utilization compared to one-to-one deployments that serve at most one request at a time.
Furthermore, centralized serving systems eliminate the need to equip each robot with dedicated accelerators, reducing the total hardware footprint and capital cost, especially since not all robots måay be active simultaneously, which would otherwise lead to significant resource underutilization.

\end{itemize}

\begin{figure}[t]
  \centering
  \includegraphics[width=\linewidth]{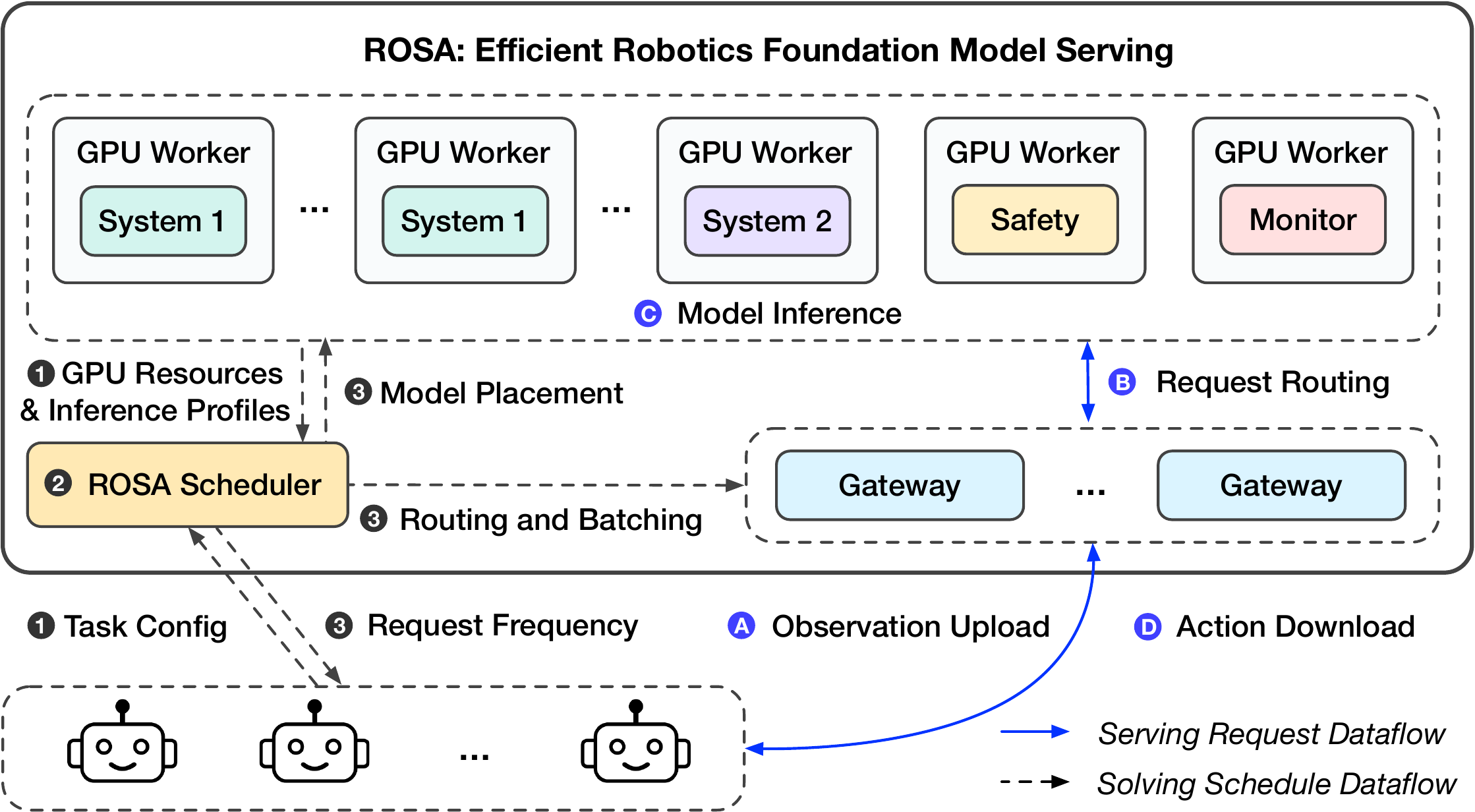}
  \vspace{-1.5em}
  \caption{\ours system overview.}
  \vspace{-1em}
  \label{fig:system_overview}
\end{figure}

\niparagraph{Principle 2: robotics-aware programming abstraction and system design.}
Robotic serving infrastructure must not only execute inference requests, but also expose a programming abstraction that specifies all robot serving requirements (R1--R3):

\begin{itemize}[wide=0pt,topsep=4pt, itemsep=4pt, parsep=0pt, partopsep=0pt]

    \item[(R1)] \textit{Orchestration across heterogeneous models.}
    Depending on the type of the task, a robot may require not only an action prediction model, but also a reasoning model for long-horizon task decomposition, a safety model to avoid collisions, and a monitor model to evaluate task progression. 

    \item[(R2)] \textit{Heterogeneous performance requirements.}
    Different robots, tasks, and operating environments impose different performance demands: some stationary robots may tolerate higher inference latency, while robots operating in dynamic environments require tighter response time.

    \item[(R3)] \textit{Safety and task-failure handling.}
    The system must let users specify how robots respond when a safety model raises a warning, an inference request misses its latency SLO, or a task-progression monitor detects task failure.
    These exceptions should trigger configured fallback actions, such as stopping execution, returning to a default safe state, resending a request, retrying the task, or escalating to human intervention.

\end{itemize}

\niparagraph{Principle 3: factory-objective-driven system scheduling.}
Instead of minimizing inference latency for a single robot, the serving scheduler should optimize a factory-level objective.
In this paper, we use \textit{weighted robot action throughput}: each robot action represents one unit of task progress, and each task is assigned a weight that captures its priority or economic value.
The serving scheduler maximizes the weighted action rate across the robot fleet\footnote{In this paper, the term \textit{fleet} refers to the set of robots rather than servers.} by deciding how to allocate servers, choose batching configurations, and route robot requests, while satisfying each task's latency and invocation-rate SLOs.

\niparagraph{Design and implementation.}
To materialize these principles, we build \ours, a RFM serving system for robot factories, as shown in \Cref{fig:system_overview}.
To plan for a serving schedule, the scheduler takesa as input (1) a declarative task configuration that specifies the robot fleet, model components, prompts, SLOs, and fallback policies, and (2) the available GPU resources and profiling information for each model~\ballnumber{1}.
The \ours scheduler combines heuristics and Integar Linear Programming (ILP) to maximize productivity while meeting performance SLOs~\ballnumber{2} and generates a schedule including model placement, request routing, batching configuration, and per-task action rates~\ballnumber{3}.
In the serving path, robot clients upload observations to a gateway~\blueballnumber{A}, which routes each model request to the scheduled GPU worker~\blueballnumber{B}; the worker executes the corresponding model inference~\blueballnumber{C}, such as System~1, System~2, safety, or monitor, and returns the result to the robot~\blueballnumber{D}.
\ours uses Ray Serve~\cite{moritz2018ray} as the distributed serving orchestrator and supports various model-specific inference runtime such as vLLM~\cite{kwon2023efficient}, PyTorch, or JAX. 

\niparagraph{Evaluation.}
We evaluate \ours on both a real Franka Panda robot and synthetic multi-robot workloads replaying real robot observations.
On eight NVIDIA H200 GPUs with up to 64 virtual robots, \ours improves productivity by up to 12.06$\times$ over dedicated serving baselines that allocate either one GPU per robot or one GPU per model per robot.
Even compared with shared-server baselines that use the same serving infrastructure but do not use the \ours scheduler, \ours improves SLO-qualified factory action throughput by up to 2.44$\times$.
Ablations show that these gains come from scheduler decisions including request-rate control, resource allocation across model components, and profiling-guided batching.

In summary, our work presents an early effort toward efficient RFM serving and makes three main \textbf{contributions}:

\begin{itemize}[left=0pt,topsep=4pt, itemsep=4pt, parsep=0pt, partopsep=0pt]

    \item  We identify two common misconceptions in existing RFM serving, and propose three design principles for robot factories: shared server-scale infrastructure, robotics-aware programming abstraction and system design, and factory-objective-driven scheduling.

    \item We design and implement \ours, a server-scale, model-agnostic RFM serving system shared across robots. \ours provides (1) a declarative task interface for specifying model composition, SLOs, safety fallbacks, and retry policies, and (2) a scheduler that maximizes factory action throughput while satisfying performance constraints.

    \item We evaluate \ours on both real robots and large-scale simulated workloads, showing that it can satisfy various robotics serving requirements while improving serving efficiency.

\end{itemize}

\section{Background and Related Work}
\label{sec:background}

\subsection{Robot Factory}
\label{sec:background:robot_factory}

The emergence of robotics foundation models (RFM) enables a new paradigm of general-purpose robots (e.g., humanoid or arms) to perform a wide range of tasks, such as packaging, sorting, inspection, and assembly assistance~\cite{figure2025bmwdeployment, vaziri2026teslaoptimusfremont, park2025vulcan, agaskar2025deepfleet}.
A promising near-term deployment scenario is the \textit{robot factory}: a human-supervised industrial environment in which fleets of general-purpose robots execute known production tasks while human operators handle exceptions.
In such a factory, robots may either work on uniform tasks, such as bulk item packaging or clothes folding, or be orchestrated across heterogeneous tasks, much like human workers handling different tasks in an assembly line.
Early versions of this robot factory vision are already being explored by industry leaders.
For example, Figure AI has deployed humanoid robots in BMW manufacturing facilities for sheet-metal handling and other automotive production tasks.
Amazon has scaled large fleets of warehouse robots for inventory movement, sortation, and package handling, with human operators available to intervene when robots fail to complete an action, encounter uncertainty, or reach a safety-critical state.

\subsection{Robotics Foundation Models}
\label{sec:background:rfm}

\subsubsection{Action Models}
\label{sec:background:rfm:action_models}

\niparagraph{System 1 for low-level action generation.}
Action models are the core building block of RFMs: they take robot observations and task instructions as input and generate sequences of low-level actions for the robot to execute.
For example, Vision-Language-Action (VLA) models~\cite{zhao2023learning, black2024pi0, kim2024openvla, zitkovich2023rt, bjorck2025gr00t} typically combine visual perception and semantic language understanding through a vision-language model (VLM) backbone, then decode robot actions using an action-generation head such as a diffusion transformer (DiT).
World Action Models (WAMs)~\cite{ye2026world, agarwal2026cosmos, zhang2026qwen}, by contrast, predict actions together with future environment states, such as upcoming camera observations, enabling the model to reason about both what the robot should do to reach the desired future state.
These action models are often viewed as the robot's ``System 1'': they provide fast, reactive control for translating simple instructions into executable actions.

\subsubsection{Multi-Model Pipelines Beyond Action Models}
\label{sec:background:rfm:multi_model_pipelines}
In a robot factory, serving a low-level action model is often not enough.
Reliable robot execution requires a multi-model pipeline that combines action generation with higher-level planning, operational safety, and task progression monitoring.

\niparagraph{System 2 for high-level planning.}
Complex real-world tasks require not only reactive control but also deliberative ``System 2'' reasoning, analogous to slow thinking in contrast to the fast reactive behavior of System 1~\cite{bjorck2026vesta, figure2025helix, zhang2024hirt, song2025hume}.
These System 2 models, often VLMs with reasoning capabilities, decompose long-horizon tasks into simpler subgoals and pass the generated language instructions to the System 1 model for action generation.
For example, in an assembly-kit task, a System 2 model may inspect the workspace, identify the required parts, decide an ordering such as ``pick the screw pack, place it in the tray, then insert the manual,'' and issue each subgoal to the action model for execution.
Because System 2 operates at the task-planning level, it typically runs at a lower frequency than System 1, such as once every several or tens of System 1 invocations.

\niparagraph{Operational safety.}
Robots must also satisfy safety constraints, such as stopping before colliding with nearby humans, objects, or workspace boundaries.
One way to support this is to invoke a safety VLM that takes the current camera observation and a safety-constraint prompt as input, and judges whether the robot's current state or planned action is unsafe~\cite{khan2025safety, munguia2026chemist, wang2026probing}.
Such models can be invoked periodically, for example once per second, to provide higher-level semantic safety checks alongside low-level onboard safety mechanisms.

\niparagraph{Task progression monitoring.}
Robot factories also need mechanisms to determine whether a task has succeeded or failed, as existing action models typically only predict actions without explicitly verifying task completion~\cite{duan2024aha, luo2024vision, elmallah2025score}.
A task-monitor VLM judge, similar to the safety model, can periodically inspect robot observations and execution history to assess task progress.
When a failure is detected, the system can trigger a retry, update the System 2 prompt with past execution feedback, or escalate to a human operator after repeated failures.

\subsection{Related Work and Research Gap}

Existing work on efficient RFM inference has focused predominantly on System 1 action models.
These efforts reduce computation through smaller models~\cite{wen2025tinyvla, shukor2025smolvla, lin2025evo, sun2026dadu}, quantization~\cite{kim2024openvla, wang2025bitvla}, selective VLM layer skipping~\cite{yue2024deer, yangdysl}, fewer denoising steps in diffusion-based action experts~\cite{bjorck2025gr00t}, KV-cache reuse for multi-token action prediction~\cite{kim2025fine}, and dynamically adjusting the reasoning model invocation frequency~\cite{liu2026should}.
Inference latency can also be reduced through system-level optimizations such as CUDA graphs and operator fusion~\cite{ma2025running}.

While these works are important steps forward, they are still not sufficient for efficient RFM serving in robot factories.
First, they primarily optimize System 1 action prediction, which is only one component of the multi-model pipeline required for reliable robot operation (\S\ref{sec:background:rfm:multi_model_pipelines} and \S\ref{sec:solution:problem_formulation}).
Second, they typically assume a single inference system serving a single robot, which does not capture factory-scale deployments with fleets of robots potentially sharing a pool of compute resources (\S\ref{sec:background:robot_factory} and \S\ref{sec:solution:cluster}).
Third, as a consequence, existing systems have not provided the programming interface and scheduling mechanisms needed for multi-robot, server-scale serving of multi-model RFM pipelines (\S\ref{sec:solution:abstraction} and \S\ref{sec:scheduler}).
Finally, many techniques are tied to specific model architectures, making it unclear how they can be adapted to rapidly evolving RFM designs; for example, RFM architectures have shifted from autoregressive VLA models to diffusion-based VLA models and emerging world-action models in only the past two years~\cite{kim2024openvla, black2024pi0, bjorck2025gr00t, ye2026world}. 
In this paper, we address these gaps.

\section{\ours: Efficient Serving for Robot Factories}
\label{sec:solution}

We present \ours, a server-scale, GPU-pool-based RFM serving system for robot factories. 
As shown in \Cref{fig:system_overview}, a fleet of robots sends inference requests to a shared GPU cluster, which uses \ours scheduler to determine optimal serving strategies.
In this section, we formalize the robot-factory serving problem, motivate centralized cluster-scale serving, introduce the declarative programming abstraction, present \ours scheduling mechanisms, and finally describe the system implementation.

\subsection{RFM Serving Problem Formulation}
\label{sec:solution:problem_formulation}

\niparagraph{Robot factory serving scenario.}
A robot factory consists of a fleet of robots controlled by a shared set of robotics foundation models.
These robots may perform either the same task or different tasks, and each task may require a different composition of RFMs, such as System 1 action models, System 2 planning models, safety models, and task-monitor models.
Each task also specifies its own prompts and performance requirements, including latency and throughput targets.
In this setting, the goal of the RFM serving system is twofold.

\niparagraph{Basic serving requirements.}
First, the system must satisfy the serving requirements of each robot by (1) deploying the required model composition and (2) meeting the associated performance constraints.
For example, a simple pick-and-place task may require only a System 1 action model and a task-monitor model, while a long-horizon assembly task may additionally require periodic System 2 planning as well as a safety model that runs at 2~Hz with a 500~ms latency constraint.

\niparagraph{Optimization objectives.}
Second, the system should optimize for factory-level objectives rather than simply minimizing the latency of individual inference requests.
In many robot-factory settings, reducing latency below a task-specific threshold may provide little additional benefit to task success, while requiring excessive compute resources, such as dedicating one or more GPUs to a single robot.
Instead, natural factory-level serving should maximize factory throughput, measured as the total weighted robot action rate across the robot fleet, while satisfying each task's latency and throughput requirements.

\subsection{Shared RFM Serving Architecture}
\label{sec:solution:cluster}

To meet the system goals specified in \S\ref{sec:solution:problem_formulation}, we argue that robot factories favor a \textit{centralized, server-scale RFM serving system} shared across a fleet of robots, while the robot-side SoC only needs minimal compute capacity to handle local control and safety fallback operations.
Compared with one-to-one deployments, where each robot is paired with onboard accelerators such as NVIDIA Jetson-class SoCs or a nearby GPU workstation or edge server, a pooled serving system provides \textbf{three key advantages}:

\begin{itemize}[wide=0pt, topsep=4pt, itemsep=4pt, parsep=0pt, partopsep=0pt]

\item \textit{Support for larger models and faster inference.}
Onboard accelerators like Jetson are substantially more constrained than data-center GPUs in compute throughput, memory capacity, and memory bandwidth.
As RFMs grow in size and complexity, some models, e.g., reasoning models, may be difficult or impossible to serve efficiently on robot-local hardware.
Centralized clusters can instead use a pool of data-center-class accelerators, enabling larger models and lower inference latency.

\item \textit{Longer operational duration for mobile robots.}
For battery-powered mobile robots, onboard inference can consume a significant fraction of the robot's power budget and reduce operating time, e.g., by up to roughly half for Figure AI's Figure~02 robot.
Offloading RFM inference to centralized infrastructure reduces onboard compute and thermal demands, extending battery duration and improving robot availability.
The robot can still retain lightweight local compute, such as CPUs or microcontrollers, for low-level control, emergency stopping, and other safety-critical mechanisms. 

\item \textit{Better hardware utilization and cost efficiency.}
A dedicated inference system suffers from two sources of resource under-utilization: (1) action-model inference is interleaved with physical robot execution, leaving the serving hardware idle while the robot executes an action if no other model component is running, and (2) a single robot sends only one inference request at a time, which further under-utilizes the inference hardware due to the lack of request batching.
In contrast, a shared serving system eliminates the need to provision dedicated accelerators for every robot and improves utilization by (1) serving requests from other robots while one robot is executing an action, and (2) enabling inter-robot request batching.

\end{itemize}

\niparagraph{Design trade-offs.}
Centralized serving introduces network and queuing overheads relative to dedicated serving, but these overheads are manageable in robot-factory environments.
First, the added network latency can be kept small by provisioning stable wired or wireless networks, whose cost is modest compared with GPU infrastructure.
For example, well-engineered WiFi~7 deployments can achieve sub-5~ms latency~\cite{liu2023first, alsakati2023performance}, which is one to two orders of magnitude smaller than the RFM inference latencies we observe, e.g., tens to hundreds of milliseconds for action models and multiple seconds for System~2 reasoning.
Second, sharing resources across robots can introduce queuing delay, but this can be controlled through scheduling.
\ours uses calibrated scheduling to ensure that robot performance requirements are met while improving overall factory-level efficiency, as we show in \S\ref{sec:scheduler}.

\niparagraph{Robot-side system responsibilities.}
Centralized serving does not completely eliminate the need for onboard SoCs, which retain two main responsibilities.

\textit{First, the robot-side system performs high-frequency local control.}
The server-side action model outputs an action chunk that describes a trajectory or motion plan, but this output must still be converted into actuator-level commands, such as joint torques, using real-time proprioceptive and force feedback.
This control loop often runs at 100~Hz or above with a small neural network~\cite{luo2025sonic, li2025amo}.
Due to the tight latency budget for network communication, the control policy computation should remain on the robot, which is feasible as prior work such as SONIC~\cite{luo2025sonic} shows that such policies can be small enough to run on CPU-based onboard compute for humanoid robots.

\textit{Second, the robot-side runtime enforces local safety and failure handling.}
When a cluster-served safety model raises a warning, the robot must be able to react locally.
In addition, each model may have a latency SLO, and the robot-side runtime must detect missed inference deadlines caused by network jitter or server failures.
Upon such safety or timing violations, the robot transitions to a safe state specified by our programming interface (\S\ref{sec:solution:abstraction}), such as stopping execution or planning a trajectory back to the robot’s default safe state.

\subsection{\ours Programming Abstraction}
\label{sec:solution:abstraction}

To capture the performance and safety requirements specified in \S\ref{sec:solution:problem_formulation}, \ours provides a declarative programming interface for describing the factory serving workload.
\Cref{fig:config_api} shows an example of this interface.
The descriptor first declares the server cluster and robot fleet used by the deployment (lines~1--12).
It then specifies the robot tasks to be served, where each task descriptor includes attributes including model composition, performance requirements, safety violation handling, and task retry policy. We now elaborate these attributes.

\begin{figure*}[t]
\centering

\begin{minipage}[t]{0.48\textwidth}
\vspace{-0.4em}
\begin{Verbatim}[fontsize=\scriptsize,breaklines=true,frame=single,numbers=left,numbersep=4pt]
# Part 1: Server cluster
server_cluster:
  num_servers: 8
  gpu_type: NVIDIA H200

# Part 2: Robot fleet
robot_fleet:
  - task: pick_and_place_simple
    num_robots: 16

  - task: inspect_product
    num_robots: 16

# Part 3.1: Task I (Pick and Place)
pick_and_place_simple:
  pipeline:
    action_period_ms: 200

  task_retry:
    max_task_retries: 3
    on_max_task_retries: stop_and_call_human

  safety_and_slo_violation:
    max_consecutive_safety_replan: 10
    max_consecutive_slo_violation: 3
    on_max_violation: stop_and_call_human

  components:
    system1:
      model: nvidia/GR00T-N1.6-3B
      prompt: "pick package and place in bin"
      slo_ms: 200
        fallback: stop_and_resend

    monitor:
      model: Qwen2.5-VL-7B-Instruct
      prompt: "ongoing, done, or failed?"
      freq_hz: 0.5
      slo_ms: 2000
        fallback: stop_and_resend
\end{Verbatim}
\end{minipage}
\hfill
\begin{minipage}[t]{0.48\textwidth}
\vspace{-0.4em}
\begin{Verbatim}[fontsize=\scriptsize,breaklines=true,frame=single,numbers=left,firstnumber=last,numbersep=4pt]
# Part 3.2: Task II (Inspect Product)
inspect_product:
  pipeline:
    action_period_ms: 500
    system2_to_system1_call_ratio: 1

  task_retry:
    max_task_retries: 3
    on_max_task_retries: stop_and_call_human

  safety_and_slo_violation:
    max_consecutive_safety_replan: 10
    max_consecutive_slo_violation: 3
    on_max_violation: stop_and_call_human

  components:
    system1:
      model: nvidia/GR00T-N1.6-3B
      prompt: "move to inspect package face"
      slo_ms: 500
        fallback: stop_and_resend

    system2:
      model: Qwen2.5-VL-7B-Instruct
      prompt: "check defects; choose next view"
      slo_ms: 2000
        fallback: use_last_plan

    safety:
      model: Qwen2.5-VL-3B-Instruct
      prompt: "is the workcell safe?"
      freq_hz: 2
      slo_ms: 500
        fallback: stop_and_replan

    monitor:
      model: Qwen2.5-VL-7B-Instruct
      prompt: "ongoing, done, or failed?"
      freq_hz: 0.5
      slo_ms: 2000
        fallback: stop_and_resend
\end{Verbatim}
\end{minipage}

\vspace{-0.5em}
\caption{\ours provides a declarative programming abstraction specifies the servers, the robot fleet, and detailed task requirements.}
\label{fig:config_api}
\end{figure*}

\niparagraph{Model composition.}
A task descriptor first specifies the set of model components required by the task.
For example, the simple pick-and-place task (line~15) uses only two components: a System~1 action model and a task-progression monitor.
In contrast, product inspection (line~42) requires a more complex pipeline, adding a System~2 planner and a safety checker.

The descriptor also specifies the dependencies among model components.
For example, the inspection task couples System~2 planning, System~1 action generation, and physical action execution.
It invokes System~2 once for every System~1 action prediction (line~45), allowing the planner to re-evaluate the product after each small motion.
Each System~1 prediction is then followed by 500~ms of action execution (line~44).
In contrast, the safety and task-progression monitor components are independent periodic checks, whose inference requests are triggered independently of the System~1--System~2 chain.

\niparagraph{Performance requirements.}
Each model component in a task may declare its own P99 latency requirement and invocation rate.
For example, the pick-and-place task requires each System~1 action prediction to complete within 200~ms (line~17), lower than that of the inspection task (line~60).
As another example, the safety-checking model in the inspection task is invoked twice per second and has a P99 latency requirement of 500~ms (line~72).

\niparagraph{Safety and SLO violations handling.}
Safety violations can arise either because a safety model explicitly raises a warning or because an inference request misses its latency SLO, making the returned prediction stale.

When the safety model raises a warning, such as detecting a human near the workcell, the robot executes the configured safety fallback, e.g., stopping and replanning (line~74).
An alternative option is planning a trajectory back to the robot’s default safe state instead of just stopping the execution.
If repeated replanning cannot resolve the safety concern, the task-level retry policy escalates to human intervention (line~54).

SLO violations are handled separately.
Although the serving system aims to satisfy latency SLOs with high probability, e.g., 99\%, violations may still occur due to network jitter, server overload, or network and serving failures.
When a component misses its SLO, the robot executes the component-level fallback policy, such as stopping the robot and resending the request (line~61).
If several consecutive SLO violations occur, indicating that the server may be unavailable or overloaded, the robot escalates to human intervention according to the configured retry policy (line~54).

\niparagraph{Task retry policy.}
A robot may fail at the task level, for example by failing to place an object into the target bin.
Such failures are detected by the task-progression monitor, which classifies the task state as ongoing, done, or failed (line~78).
When a task failure is detected, the system resets the robot to its initial state and retries the task.
If the task continues to fail after the configured retry budget, e.g., three retries, the system halts execution and requests human intervention (line~48).

\subsection{\ours Scheduler}
\label{sec:scheduler}

\subsubsection{Scheduler Overview}
\label{sec:scheduler:formulation}

\niparagraph{Scheduling decisions.}
The \ours scheduler turns the declarative workload specification in \Cref{fig:config_api} into a serving schedule with four main decisions:
(1) \emph{placement}: which model component each server hosts;
(2) \emph{routing}: which server each robot uses for each model request; 
(3) \emph{batching}: which batching configuration each server uses; and
(4) \emph{request rate}: which inference request frequency each robot uses for each model.

\begin{figure}[t]
  \centering
  \includegraphics[width=\linewidth]{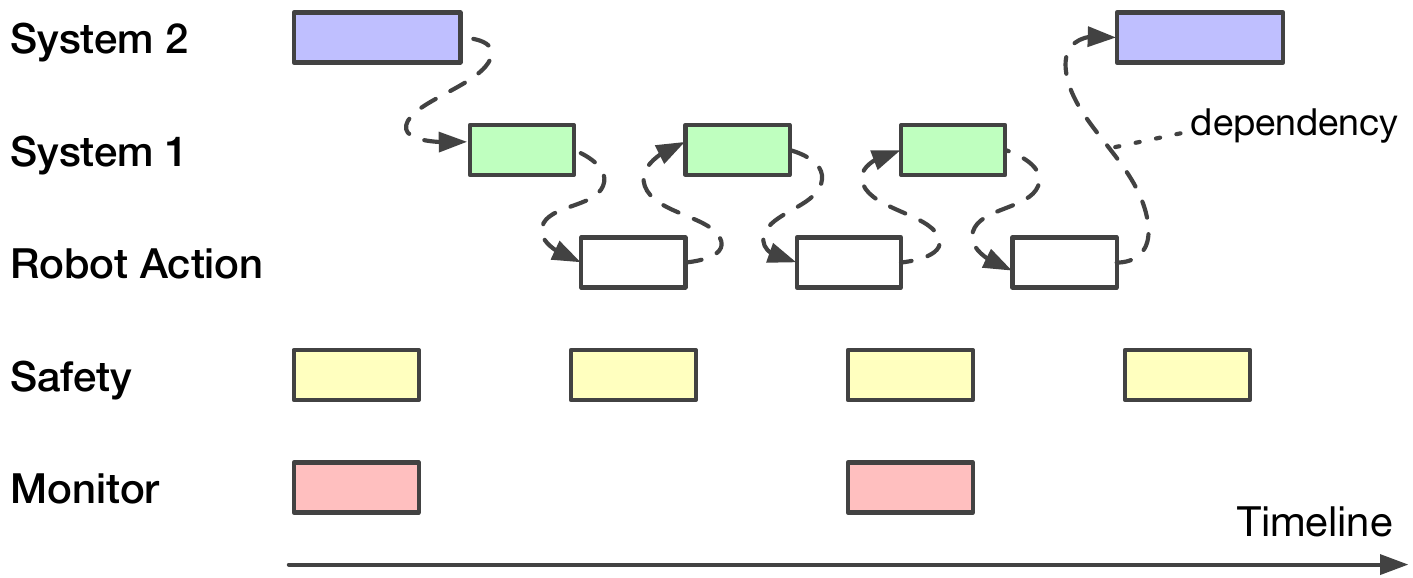}
  \vspace{-1.5em}
  \caption{An example multi-model RFM inference pipeline.}
  \vspace{-1em}
  \label{fig:timeline}
\end{figure}

Since \ours targets robot-factory deployments, the scheduler makes several workload and serving assumptions.
First, the expected workload is known before serving starts: the number of robots, task specifications, available GPUs, per-component SLOs, and the constant system1/system2 invocation ratio are provided by the declarative configuration.
This allows \ours to compute a static serving schedule that maximizes factory throughput for long-running workloads.
Second, given the known workloads, the scheduler can leverage the profiling data for each model.
Third, we assume typical synchronous execution, where System~1 inference is interleaved with action execution (\Cref{fig:timeline}) rather than overlapped with it, because asynchronous execution can reduce task accuracy when predictions are generated from stale observations~\cite{agouzoul2026understanding, black2025real, sendai2025leave, tang2025vlash}.
Fourth, as many robots share the GPU pool, each model receives enough aggregate requests, thus \ours dedicates each GPU worker to a single model rather than colocating multiple models, avoiding performance interference that can lead to SLO violations.
Finally, for batching, \ours searches over various batch sizes for System~1 action models that satisfy the latency SLO. 
For other autoregressive generation models, \ours assumes continuous batching, where individual requests are directly added to and removed from ongoing inference batches, following the convention of existing LLM serving systems~\cite{christoph2025orca,kwon2023efficient}.

\niparagraph{Scheduling requirements and objectives.}
Let $R$, $S$, and $M$ be the sets of robots, servers, and model components, respectively.
Each robot $r\in R$ requires a set of model components $M_r\subseteq M$.
For each required component $m\in M_r$, the task specification may provide a latency deadline $D_{r,m}$ and, for periodically invoked components, a fixed invocation rate $\bar{\lambda}_{r,m}$.
A schedule is feasible only if every required component is served and all specified latency and invocation-rate requirements are satisfied.

Among feasible schedules, \ours optimizes factory-level productivity rather than individual request latency.
For a homogeneous fleet, all robots execute the same task and have the same value weight, so the scheduler searches for a shared action rate $f$, where $f_r=f$ for all robots $r\in R$, and maximizes factory action throughput, i.e., $\max \sum_{r\in R} f_r$.
For a heterogeneous fleet, robots are grouped by task class $c$; class $c$ contains $K_c$ robots, receives action rate $f_c$, and has value weight $v_c$.
\begin{equation}
    \max_{\mathbf{f}} \sum_{c=1}^{C} v_c K_c f_c
    \quad
    \textrm{s.t.}\quad
    K_c f_c \geq F_c^{\min},\ \forall c .
    \label{eq:scheduler-hetero-objective}
\end{equation}
Here $F_c^{\min}$ is the minimum total action throughput required by task
class $c$, which prevents the optimizer from starving low-weight or
high-cost task classes.

To optimize this objective, the scheduler classifies model components into two groups.
As illustrated by the inference timeline in \Cref{fig:timeline}, \emph{goal-coupled models}, $\mathcal{G}=\{\mathrm{S1},\mathrm{S2}\}$, determine the robot's action rate $f$.
In contrast, \emph{obligation models}, $\mathcal{O}=\{\mathrm{safe},\mathrm{mon}\}$, run at configured frequencies and must satisfy latency SLOs, but serving them faster than required does not directly improve the factory objective.
The scheduler therefore provisions obligation models with the minimum feasible server footprint, then uses the remaining resources to maximize the rates of goal-coupled models.

\niparagraph{Steady-state scheduling and runtime adaptation.}
Factory workloads are typically long running, and the number of robots assigned to each task is often known in advance.
\ours therefore uses a steady-state solver to compute high-quality schedules for the expected workload.
This solver runs at cluster-management timescales and can afford profiling-driven feasibility checks and ILP-based packing.
Runtime changes, such as robot arrivals or GPU failures, are handled by faster adaptation mechanisms: placement-preserving admission control, hot-standby failover, and background global rescheduling when local adaptation is insufficient (\S\ref{sec:scheduler:runtime}). 

The rest of this section first introduces the steady-state solver for robot fleets serving homogeneous and heterogeneous tasks~(\S\ref{sec:scheduler:homogeneous} and \S\ref{sec:scheduler:heterogeneous}), and then describes the runtime adaptation mechanisms~(\S\ref{sec:scheduler:runtime}).

\subsubsection{Homogeneous Task Scheduling}
\label{sec:scheduler:homogeneous}

We first consider a homogeneous fleet in which all robots share the same task configuration: the same required models, action execution time $t_{act}$, System~2 horizon $H$, and latency requirements.
In this case, the objective in \S\ref{sec:scheduler:formulation} reduces to maximizing a single shared action rate $f$.

\vspace{1em}
\begin{adjustbox}{scale=0.75}
\begin{minipage}{1.33\linewidth}
\begin{Verbatim}[fontsize=\scriptsize,breaklines=true,frame=single]
# Algorithm: homogeneous task scheduling
# Step 1: provision obligation models
pi_obl, S_obl = provision_obligations(R, S, O, L)
# Step 2: binary search for maximum action rate
S_goal = S - S_obl
lo, hi, best = 0, f_max, None
while hi - lo > eps:
    f = (lo + hi) / 2 
    # Step 2.1: generate feasible per-server configs
    P = enumerate_server_configurations(f, L, S_goal)
    # Step 2.2: solve full-system configs
    pi_goal = pack_configurations_ILP(P, R, S_goal)
    # Step 2.3: check frequency feasibility
    if pi_goal is feasible and f <= closed_loop_rate(pi_goal):
        best = (pi_obl + pi_goal, f); lo = f
    else:
        hi = f
return best
\end{Verbatim}
\end{minipage}
\end{adjustbox}

\niparagraph{Step 1: provision obligation models.}
The scheduler first computes the aggregate load imposed by obligation models:
\begin{equation}
    \lambda_{\mathrm{safe}} = \sum_{r\in R} \bar{\lambda}_{r,\mathrm{safe}},
    \qquad
    \lambda_{\mathrm{mon}} = \sum_{r\in R} \bar{\lambda}_{r,\mathrm{mon}}.
\end{equation}
It then finds the minimum feasible server footprint for these models using measured serving profiles.
For each model $m$ and hardware type $h$, \ours measures a latency distribution $L_{m,h}(b,\lambda)$ for each batching configuration $b$ and aggregate request rate $\lambda$.
For example, for each System~1 batch size, we profile latency at ten request rates, ranging from 10\% to 100\% of the measured maximum throughput for that batch size.
A server assignment for an obligation model is feasible only if the required SLO percentile of this distribution is below the model's latency deadline.

\niparagraph{Step 2: search for the maximum action rate.}
After provisioning obligation models with minimal resources, the scheduler uses the remaining servers $S_{\mathrm{goal}}$ to maximize the shared action frequency $f$.
It performs a binary search over $f$, between zero and an upper bound $f_{\max}$ computed by assuming zero inference latency.
For each candidate $f$, each robot induces request rates
$\lambda_{r,\mathrm{S1}}=f$ and $\lambda_{r,\mathrm{S2}}=f/H$ for the goal-coupled models.

\niparagraph{Step 2.1: enumerate feasible per-server configurations.}
For the candidate $f$, \ours constructs a set of feasible server configuration $\mathcal{P}(f)$ from the measured serving profiles.
A configuration $p\in\mathcal{P}(f)$ specifies the model it hosts, the batch size, and the number of robots it serves.
For example, a System~1 server configuration may represent one server using batch size $b$ to serve $k$ robot streams, producing aggregate load $k f$.
A System~2 server configuration for $k$ robot streams instead sees load $k f/H$ with continuous batching ($b=1$).
A server configuration is valid only if it fits in memory and satisfies the required latency SLO under the corresponding load.
Thus, SLO feasibility is enforced during configuration generation rather than as a separate ILP constraint in the next step.

\niparagraph{Step 2.2: choose server configurations with an ILP.}
Given $\mathcal{P}(f)$, the scheduler solves an integer linear program (ILP) to cover all robot streams using the remaining servers.
Let $y_p\in\mathbb{Z}_{\geq0}$ be the number of servers assigned to configuration $p$, and let $a_{p,m}\in\mathbb{Z}_{\geq0}$ be the number of robot streams of model $m$ served by one server using configuration $p$.
The ILP solves $y_p$ with the following constraints:
\begin{equation}
\small
    \sum_{p\in\mathcal{P}(f)} y_p \leq |S_{\mathrm{goal}}|,
    \quad
    \sum_{p\in\mathcal{P}(f)} y_p a_{p,m}=|R|~(\forall m\in\mathcal{G}).
\label{eq:scheduler-configuration-pack}
\end{equation}
The first constraint limits the number of selected servers, while the second ensures that every robot is assigned to a server for every required goal-coupled model.
Among feasible packings, the ILP minimizes a tie-breaking cost based on measured latency and the number of servers used.

\niparagraph{Step 2.3: validate and update the action rate.}
If the configuration packing is feasible, the scheduler checks whether the induced latencies still support the candidate action rate.
A robot does not generate unbounded open-loop inference traffic: it waits for action-model inference, physically executes the action, and amortizes any System~2 planning latency across the actions that reuse the plan.
Let $\ell_{\mathrm{S1}}$ and $\ell_{\mathrm{S2}}$ be the measured mean latencies induced by the selected configurations.
These latencies impose a closed-loop upper bound on the achievable action rate:
\begin{equation}
    f \leq f_{\mathrm{loop}} =
    \frac{1}{
        t_{act}
        + \ell_{\mathrm{S1}}
        + \ell_{\mathrm{S2}}/H
    }
    \label{eq:scheduler-cycle}
\end{equation}
If both the packing constraints in \Cref{eq:scheduler-configuration-pack} and the closed-loop condition in \Cref{eq:scheduler-cycle} hold, the scheduler records the candidate schedule and searches for a higher $f$; otherwise, it searches for a lower $f$.

\subsubsection{Heterogeneous Task Scheduling}
\label{sec:scheduler:heterogeneous}

We next consider robot fleets that consist of different tasks, such as the example in \Cref{fig:config_api}, where some robots run a simple pick-and-place task while others run product inspection.
Different tasks may use different model components, action periods, System~2 invocation ratios, and latency SLOs.
As introduced in \S\ref{sec:scheduler:formulation}, \ours groups robots by task class $c$, where class $c$ contains $K_c$ robots and receives action rate $f_c$.
Thus, heterogeneous scheduling replaces the single homogeneous rate $f$ with a class-rate vector $\mathbf{f}=(f_1,\ldots,f_C)$.

Compared with homogeneous scheduling, heterogeneous scheduling changes two parts: the scalar binary search becomes a multi-dimensional search, and each server configuration may need to account for task classes. 
\ours handles the first with adaptive frontier search and the second with isolated packing plus compaction.

\begin{figure}[t]
  \centering
  \includegraphics[width=\linewidth]{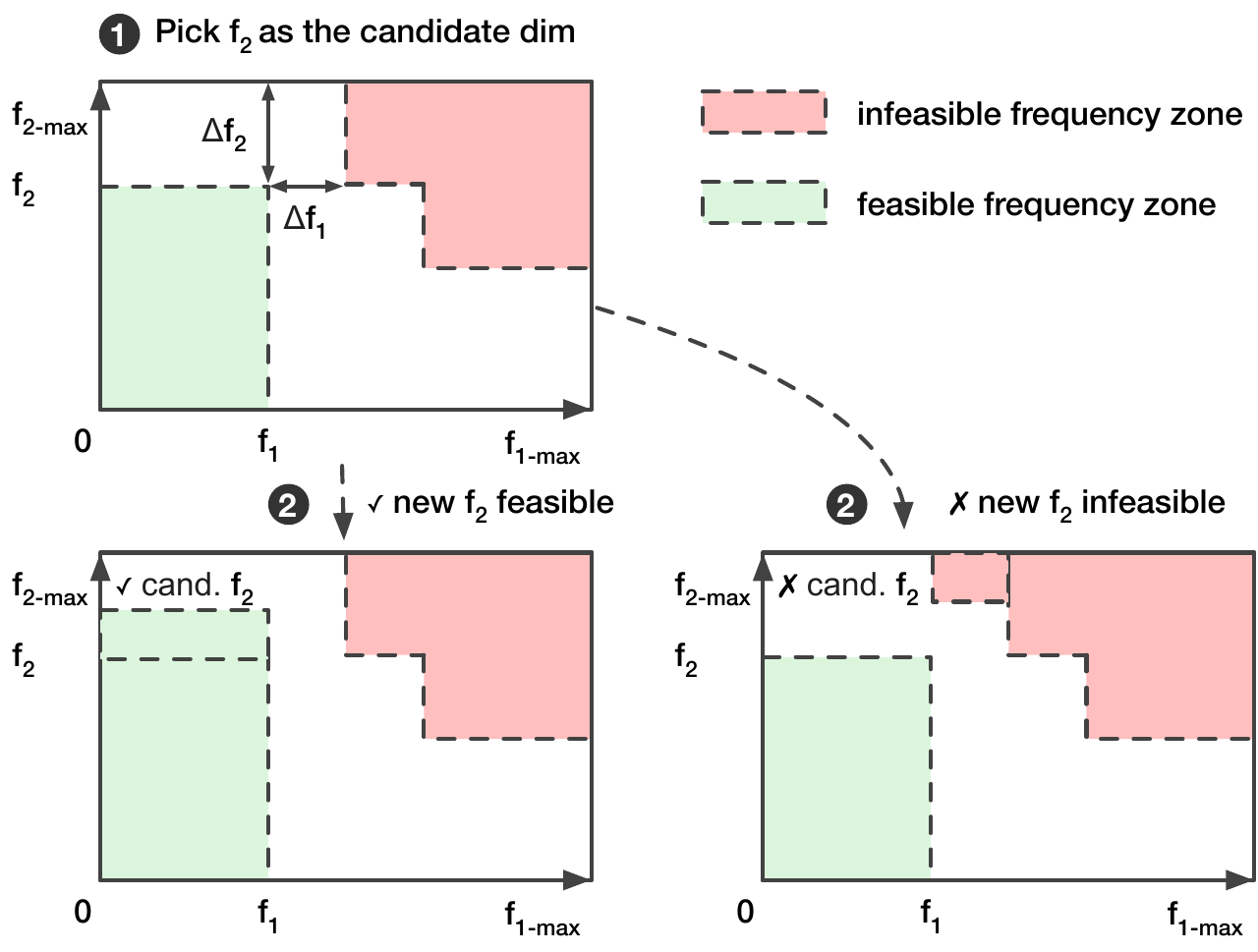}
  \vspace{-1em}
  \caption{Adaptive frequency search over heterogeneous tasks.}
  \vspace{-1em}
  \label{fig:heterogeneous_search}
\end{figure}

\niparagraph{Adaptive search over heterogeneous action rates.}
While homogeneous scheduling uses binary search over a single action rate $f$, heterogeneous scheduling must search over a rate vector $\mathbf{f}$.
A dense grid search over this multi-dimensional space has complexity exponential in the number of task classes.
\ours therefore uses a greedy adaptive frontier search to navigate the space.
\Cref{fig:heterogeneous_search} illustrates a two-task example with action rates $f_1$ and $f_2$.
The green region denotes rate vectors that are feasible under the placement and SLO checks described next, while the red region denotes rate vectors that are infeasible.
At each iteration, \ours selects the rate dimension with the largest potential weighted objective gain according to \Cref{eq:scheduler-hetero-objective}.
It then evaluates a new candidate vector by moving halfway between the current feasible rate and the optimistic upper bound along that dimension.
If the candidate vector is feasible (as we will describe in the next paragraph), \ours advances the feasible frontier; otherwise, monotonicity allows \ours to prune all rate vectors that are component-wise larger.
The search stops when the remaining possible improvement falls below a tolerance or when a time budget is reached.

\niparagraph{Frequency feasibility check with ILP-heuristic packing.}
In contrast to homogeneous scheduling, which checks action frequency feasibility with ILP, we introduce a two-stage ILP-heuristic primitive that is repeatedly used in heterogeneous scheduling.
First, \ours performs \emph{isolated packing}: each server configuration serves one model component for one task class, and an ILP selects configurations to cover the robot streams.
Second, \ours applies \emph{compaction}: servers hosting the same model component for different classes are greedily merged when doing so remains SLO-feasible.
This design avoids the combinatorial complexity of directly enumerating mixed heterogeneous configurations, where each server configuration would need to specify (1) which task classes it serves, (2) how many robots from each class it serves, and (3) the batching configuration.
By letting the ILP reason only about single-class configurations, \ours keeps configuration enumeration small.
Then, the greedy compaction step recovers cross-class sharing opportunities.
For each server, the scheduler uses the measured latency profile to estimate its SLO-compatible request-rate capacity and remaining headroom.
It tries to iteratively merge low-load servers into high-headroom servers, accepting a merge only if the combined server still satisfies the latency SLOs of all assigned classes.

This primitive is applied to each model component in the task pipeline.
As in homogeneous scheduling, \ours first provisions the minimum feasible server footprint for obligation models.
The remaining servers are then used for the goal-coupled System~1 and System~2 models that determine the action rate.
For each candidate rate vector $\mathbf{f}$, \ours first applies isolated ILP packing to each goal-coupled model.
If the resulting placement uses more than the available goal-server budget, \ours applies greedy compaction to determine whether cross-class sharing can reduce the placement to fit within the server budget.

\subsubsection{Admission Control and Fault Tolerance}
\label{sec:scheduler:runtime}

\ours supports two paths for admitting new robots into the serving system.
The first path performs \emph{placement-preserving admission}.
Here, the scheduler keeps the current model placement fixed and checks whether the new robot can be routed to existing servers, possibly with updated batching and per-robot action rates, without violating the SLOs of either the new robot or the existing robots sharing those servers.
If this check succeeds, \ours admits the robot and updates only the affected routing and rate assignments, avoiding a full system restart.
The second path performs \emph{global rescheduling}.
If placement-preserving admission fails, or if it would lead to poor efficiency, the \ours scheduler recomputes the full schedule, including model placement, routing, batching, and action rates.
This path can achieve better efficiency but may require model migration and coordinated routing updates, causing higher overhead.

When failure tolerance is enabled, \ours uses hot-standby GPU nodes.
The scheduler reserves standby GPUs and keeps the corresponding model weights loaded, but excludes these GPUs from the normal serving routes.
When an active serving GPU fails, the router simply redirects the affected robot streams to the corresponding warm standby GPU, without searching for a new schedule.
An alternative design is active-active redundancy, where backup capacity also serves normal traffic and failures are handled by rerouting requests to the remaining active GPUs.
We do not use this design because each failure changes the feasible schedules under a reduced server set.
Handling this online would require reinvoking the schedule solver and potentially updating model placement, which is both time-consuming and disruptive to robots that are otherwise unaffected by the failure.

\subsection{\ours Implementation}
\label{sec:implementation}

\ours uses Ray Serve~\cite{moritz2018ray} to manage distributed serving.
Each node hosts a gateway that receives inference requests from robot clients and routes them to the model replicas selected by the scheduler.
Ray is also responsible for launching, monitoring, and restarting the per-model serving runtimes on the assigned servers.
The runtime backend is model-dependent: autoregressive VLM components, such as System~2, safety, and monitor models, are served with vLLM, while System~1 action models are served with PyTorch or JAX depending on the model family.
This design keeps the serving runtime modular: adding a new model backend only requires wrapping it as a Ray-managed service.
The scheduler is implemented in Python and uses OR-Tools to solve the ILP subproblems.

\begin{table*}[t]
    \centering
    \setlength\dashlinedash{0.2pt}
    \setlength\dashlinegap{1.5pt}
    \setlength\arrayrulewidth{0.3pt}
    \begin{footnotesize}
    \setlength{\tabcolsep}{3pt}
    \renewcommand{\arraystretch}{0.9}
    \caption{Single-task robot fleet configurations used in the homogeneous scheduling experiments. All latency SLOs are p99.}
    \vspace{-1em}
    \label{tbl:single-workloads}
    \scalebox{1.0}{
        \begin{tabular}{M{0.7cm}!{\vrule width 0.7pt}L{3.1cm}L{3.2cm}!{\vrule width 0.7pt}M{1.6cm}M{2.3cm}M{1.9cm}M{1.9cm}!{\vrule width 0.7pt}M{1.0cm}}
            \hline
            \textbf{ID} & \textbf{Task} & \textbf{Components}
            & \textbf{System~1}
            & \textbf{System~2}
            & \textbf{Safety}
            & \textbf{Monitor}
            & \textbf{Robot} \\
            & & &
            \textbf{Freq. / SLO}
            & \textbf{Freq. / SLO}
            & \textbf{Freq. / SLO}
            & \textbf{Freq. / SLO}
            & $t_{act}$
            \\ \hline

            P1 & Pick-and-place action-only
            & S1
            & -- / 200 ms
            & -- & -- & --
            & 200 ms \\ \hdashline

            P2 & Pick-and-place simple
            & S1, monitor
            & -- / 200 ms
            & -- & -- & 0.5 Hz / 2000 ms
            & 200 ms \\ \hdashline

            P3 & Pick-and-place hard
            & S1, safety, monitor
            & -- / 200 ms
            & -- & 2 Hz / 500 ms & 0.5 Hz / 2000 ms
            & 200 ms \\ \hdashline

            P4 & Assemble kit
            & S1, S2, safety, monitor
            & -- / 200 ms
            & 1 per 10 S1 / 2000 ms
            & 2 Hz / 500 ms
            & 0.5 Hz / 2000 ms
            & 200 ms \\ \hline
        \end{tabular}
    }
    \end{footnotesize}
    \vspace{-1em}
\end{table*}

\begin{figure*}[t]
    \centering
      \includegraphics[width=\linewidth]{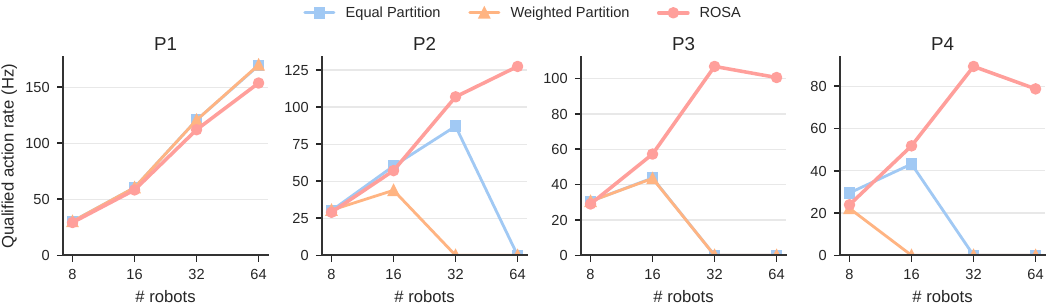}
      \vspace{-1em}
    \caption{Single-task end-to-end performance measured as SLO-qualified action throughput.}
      \vspace{-1em}
    \label{fig:e2e_performance}
  \end{figure*}

\begin{table}[t]
    \centering
    \setlength\dashlinedash{0.2pt}
    \setlength\dashlinegap{1.5pt}
    \setlength\arrayrulewidth{0.3pt}
    \begin{footnotesize}
    \setlength{\tabcolsep}{4pt}
    \renewcommand{\arraystretch}{0.9}
    \caption{Heterogeneous task workload configurations.}
    \vspace{-1em}
    \label{tbl:mixed-workloads}
    \scalebox{1.0}{
        \begin{tabular}{L{2.6cm}!{\vrule width 0.7pt}L{3.2cm}M{1.6cm}}
            \hline
            \textbf{ID} & \textbf{Task Classes} & \textbf{Robot Split} \\ \hline
            Mix2 & P2, P4 & $K/2$ each \\ \hdashline
            Mix4 & P1, P2, P3, P4 & $K/4$ each \\ \hline
        \end{tabular}
    }
    \end{footnotesize}
    \vspace{-1em}
\end{table}

\subsection{Adaptation to Future Workloads}

RFMs are evolving rapidly, so it is important to distinguish enduring system design principles from implementation choices that may change over time.
We expect the three design principles in \ours --- shared server-scale serving, a robotics-aware programming abstraction, and factory-objective-driven scheduling --- to remain broadly applicable.
However, the exact model composition and scheduling algorithm can evolve.
For example, future robot pipelines may not use exactly the four components in our current implementation, but they still need a programming abstraction that specifies model components, invocation patterns, SLOs, and fallback behaviors, as described in \S\ref{sec:solution:abstraction}.
Besides, future workloads may have more dynamic request patterns, such as adaptive System~2 invocation or irregular monitor checks, thus \ours may need a different solver strategy or a more dynamic scheduling mechanism.

\section{Evaluation}
\label{sec:evaluation}

In this section, we show that \ours significantly improves factory productivity compared to various inference systems, break down the contribution of each scheduling decision, and validate that \ours can run end-to-end on real robots.

\subsection{Experimental Setup}
\label{sec:evaluation:setup}

\niparagraph{Baseline systems:} We benchmark \ours against several baselines that use the same system implementation as \ours but run different schedules.
For dedicated serving system per robot, \textit{Baseline 1} uses one GPU per robot, and \textit{Baseline 2} uses one GPU per model required by the robot.
For shared serving architecture, \textit{Baseline 3} partitions GPUs equally across models, without an optimized batching policy, using batch size one for System~1 and continuous batching for other autoregressive models.
\textit{Baseline 4} partitions GPU resources proportional to model sizes, using the same batching policy as \textit{Baseline 3}.
All baselines issue requests as soon as a robot receives an observation, without controlling the request rate.

\niparagraph{Hardware:} We run our experiments on an 8-GPU NVIDIA H200 server.
For real robot experiments, we use the Franka Panda robot arm to validate system functionality.

\niparagraph{Workloads:} Our evaluation includes both real robot experiments and synthetic large-scale workloads.
The real robot experiments provide small-scale functional validation of the full system.
Due to the limited number of available robots, we use synthetic workloads to mimic large-scale factory deployments with tens to hundreds of robots.
These synthetic workloads replay sampled observations from the real robot setup, such as those shown in \Cref{fig:real_robot:action}.

We evaluate single-task multi-stage workloads (\Cref{tbl:single-workloads}) and mixed-task workloads (\Cref{tbl:mixed-workloads}).
The single-task workloads use pipelines of increasing complexity, from action-only System~1 execution to pipelines that include System~2, safety, and monitoring models.
The latency and throughput requirements for each model component are specified in \Cref{tbl:single-workloads}.
The mixed-task workloads combine multiple task classes in the same cluster, where different task classes may use different model pipelines and request rates.

These workloads use representative models.
System~1 uses GR00T N1.6 with PyTorch, with \texttt{torch.compile}, CUDA Graphs, and FlashAttention enabled, and also evaluates \(\pi_0.5\) with JAX. 
Both models run in their original FP16 precision to preserve accuracy.
Due to the close inference performance of the two models (\Cref{fig:s1_latency_throughput}), we use GR00T N1.6 as the default model unless otherwise specified.
System~2 reasoning model and progression monitor use Qwen-VL-7B, while safety model uses Qwen-VL-3B due to its simpler functionality.

\subsection{End-to-End Performance}
\label{sec:evaluation:e2e}

\begin{figure}[t]
  \centering
    \centering
    \includegraphics[width=\linewidth]{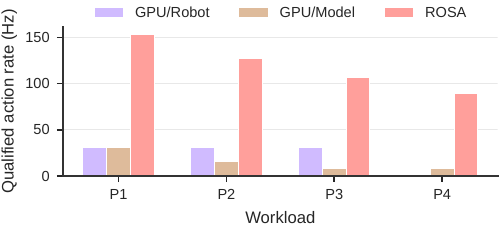}
    \vspace{-1em}
  \caption{\ours versus dedicated serving baselines.}
    \vspace{-1em}
  \label{fig:e2e_performance_vs_dedicated}
\end{figure}

We first evaluate end-to-end factory productivity, defined as the total SLO-qualified robot action throughput.
A robot action is counted only when the latency and invocation requirements introduced in \S\ref{sec:scheduler} are satisfied.
We later on break down the individual scheduling factors that contribute to this performance in \S\ref{sec:evaluation:ablation}.

\niparagraph{Single-task workloads.}
\Cref{fig:e2e_performance} and \Cref{fig:e2e_performance_vs_dedicated}  compare \ours against shared-server scheduling baselines and dedicated serving baselines on the single-task workloads described in \Cref{tbl:single-workloads}, respectively.

\textit{Compared to shared-server baselines, \ours achieves substantially higher factory productivity especially when the workload has more robots (\Cref{fig:e2e_performance}).}
The effect is clearest on P4.
With 8 robots, the baselines achieve slightly higher throughput: 29.5 qualified actions/s for equal partitioning compared with 23.9 actions/s for \ours.
This is because the GPUs are not yet saturated, so issuing requests without a rate limit does not lead to enough queueing to violate the SLO, while \ours conservatively caps the request rate using the scheduler's P99-safe prediction.
As the robot count increases to 32, however, both baselines, without scheduler-predicted rate limiting, produce zero SLO-qualified actions because their System~1 SLO meet rate drops to 0\%, whereas \ours still achieves 89.4 qualified actions/s with a 99.9\% SLO meet rate.

\textit{The benefit of \ours also grows as the pipeline has more model components when static resource allocation becomes insufficient.}
At 32 robots, \ours improves qualified throughput by 1.23$\times$ over the best shared-server baseline on P2, from 87.1 to 106.9 qualified actions/s.
For the more complex P3 and P4 pipelines at the same robot count, the shared-server baselines collapse to zero qualified throughput, while \ours sustains 106.9 and 89.4 qualified actions/s, respectively.

\textit{Compared to dedicated serving baselines, \ours achieves up to 12.06$\times$ higher factory productivity than the best dedicated configuration (\Cref{fig:e2e_performance_vs_dedicated}).}
This is due to the limited number of robots supported by these dedicated serving systems.
For example, with 8 GPUs, a one-GPU-per-robot baseline can support at most 8 robots even when multiple model components share the same GPU, while a one-GPU-per-model baseline supports only up to 2 robots for the four-component P4 pipeline.
By contrast, \ours uses the same GPU pool to serve many more robots while still satisfying latency and invocation constraints.
As a result, on P4, \ours reaches 89.4 qualified actions/s with 32 robots, while one-GPU-per-model serving reaches only 7.4 qualified actions/s with 2 robots; the one-GPU-per-robot baseline does not produce an SLO-qualified P4 operating point in our sweep.

\begin{figure}[t]
  \centering
    \includegraphics[width=\linewidth]{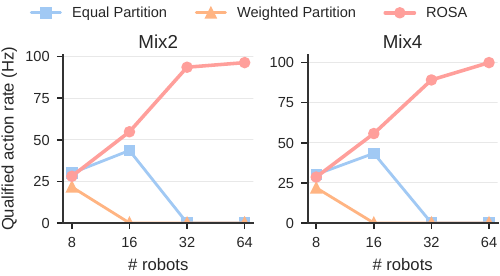}
    \vspace{-1em}
  \caption{Heterogeneous-task end-to-end performance.}
    \vspace{-1em}
  \label{fig:e2e_performance_heterogeneous}
\end{figure}

\niparagraph{Heterogeneous workloads.}
\Cref{fig:e2e_performance_heterogeneous} evaluates multiple-task workloads where robots working on different task classes share the same serving cluster.

\ours improves peak qualified throughput by 2.21$\times$ on Mix2 and 2.30$\times$ on Mix4 over the best shared-server baseline.
The shared baselines perform well only at the smallest robot count, but their qualified throughput quickly drops as the robot count increases.
This drop comes from the same two effects observed in the single-task workloads: uncapped clients continue injecting requests after bottleneck components saturate, causing SLO violations, and static resource allocation fail to allocate enough capacity to the components that limit qualified action throughput.
In heterogeneous workloads, these effects are amplified because different task classes stress different model components.
By contrast, \ours sustains high SLO compliance while allocating capacity across task classes and obligation models thanks to calculated scheduling.

\subsection{Scheduling Decision Ablation}
\label{sec:evaluation:ablation}

We next break down the sources of \ours's performance improvement.
We study four factors: (1) whether requests meet latency SLOs and therefore produce qualified action predictions, (2) the importance of controlling each robot's request send frequency, (3) the role of resource allocation across models, and (4) the effect of batching for System~1 action models.

\begin{figure}[t]
  \centering
    \includegraphics[width=\linewidth]{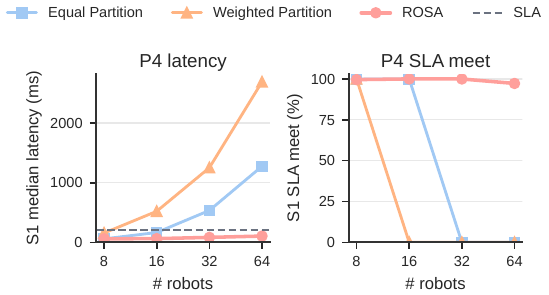}
    \vspace{-1em}
  \caption{System~1 latency and SLO qualification on P4.}
    \vspace{-1em}
  \label{fig:s1_sla_ablation}
\end{figure}

\niparagraph{SLO-qualified action prediction.}
\Cref{fig:s1_sla_ablation} compares \ours against shared-server baselines in terms of System~1 latency and SLO qualification rate on P4.
At low load, all systems satisfy the System~1 SLO.
However, as the robot count increases, the shared-server baselines continue issuing requests without controlling the request frequency, leading to significant queueing delay.
At 32 robots, equal partitioning has 0\% System~1 SLO meet rate, with median System~1 latency of 534.5~ms; weighted partitioning already reaches 0\% SLO meet rate at 16 robots, with median System~1 latency of 520.8~ms.
In contrast, \ours maintains a 99.96\% System~1 SLO meet rate at 32 robots, with median System~1 latency of 83.8~ms.
Even at 64 robots, where \ours is also under heavy load, it maintains a 97.23\% System~1 SLO meet rate instead of collapsing to zero qualified actions.

\begin{figure}[t]
  \centering
    \includegraphics[width=\linewidth]{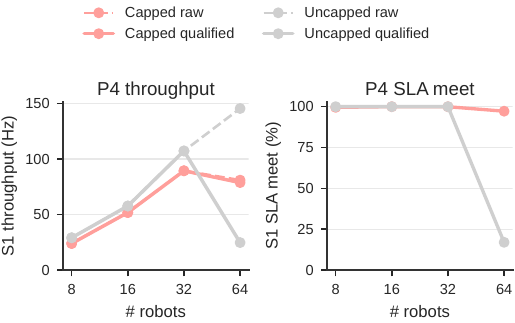}
    \vspace{-1em}
  \caption{Effect of request-rate control on P4.}
    \vspace{-1em}
  \label{fig:rate_control_ablation}
  \vspace{0.6em}
  \includegraphics[width=\linewidth]{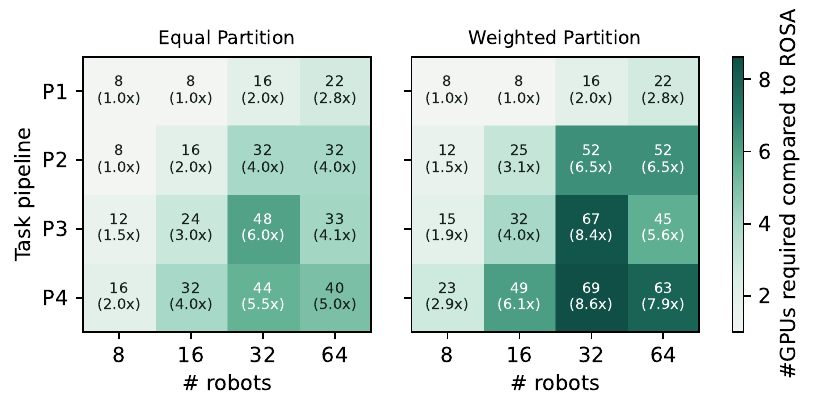}
  \vspace{-1em}
  \caption{GPU required for baseline systems to achieve \ours-level performance even when enabling request rate control.}
  \vspace{-1em}
  \label{fig:rate_control_gpu_requirement}
\end{figure}

\niparagraph{Request-rate control.}
A key part of \ours scheduling is that robots do not simply send a new System~1 or System~2 request as soon as the previous action finishes, but letting the scheduling computes an action rate to avoid overloading the server.

\Cref{fig:rate_control_ablation} shows the qualified performance and SLO meet rate for capped and uncapped request rates.
At moderate load, uncapped sending can achieve higher throughput because the serving system has not been saturated yet: at 32 robots, uncapped P4 reaches 107.3 qualified actions/s, compared with 89.4 actions/s under the capped schedule.
However, at 64 robots, uncapped sending produces 145.5 raw actions/s but only 24.8 qualified actions/s, because the System~1 SLO meet rate drops to 17.0\% under high queueing delay.
In contrast, the capped schedule maintains 78.8 qualified actions/s with a 97.2\% System~1 SLO meet rate, a 3.18$\times$ improvement in SLO-qualified throughput.

\Cref{fig:rate_control_gpu_requirement} applies \ours's request-rate control to the baseline systems and reports how many GPUs they need to match the SLO-qualified action rate achieved by \ours on eight GPUs.
According to \ours scheduler, equal and weighted partitioning baselines require up to 5.5$\times$ and 8.6$\times$ more GPUs, respectively, to match \ours, with the gap widening as the number of robots and model components increases.
This overhead comes from suboptimal resource allocation and System~1 batching decisions, which we isolate in the following ablations.

\begin{table}[t]
    \centering
    \setlength\dashlinedash{0.2pt}
    \setlength\dashlinegap{1.5pt}
    \setlength\arrayrulewidth{0.3pt}
    \begin{footnotesize}
    \setlength{\tabcolsep}{4pt}
    \renewcommand{\arraystretch}{0.9}
    \caption{Resource-allocation strategies for P4.}
    \vspace{-1em}
    \label{tbl:resource_allocation_ablation}
    \scalebox{0.98}{
        \begin{tabular}{L{2.7cm}!{\vrule width 0.7pt}M{1.2cm}M{3.6cm}}
            \hline
            \textbf{Schedule} & \textbf{Robots} & \textbf{\#GPUs S1:S2:Safe:Monitor} \\ \hline
            \ours & 8/16/32 & 5:1:1:1 \\
            \ours & 64 & 4:1:2:1 \\
            \hdashline
            Equal partition & 8/16/32/64 & 2:2:2:2 \\
            \hdashline
            Weighted partition & 8/16/32/64 & 1:3:1:3 \\
            \hline
        \end{tabular}
    }
    \end{footnotesize}
    \vspace{-1em}
\end{table}

\begin{figure}[t]
  \centering
  \includegraphics[width=.8\linewidth]{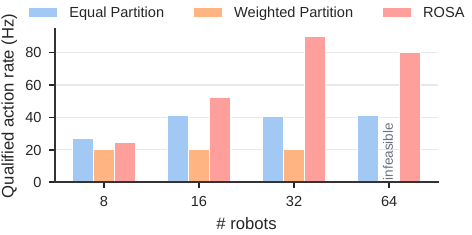}
    \vspace{-1em}
  \caption{Performance of different resource allocations on P4.}
    \vspace{-1em}
  \label{fig:allocation_ablation}
\end{figure}

\begin{figure}[t]
  \centering
  \begin{subfigure}[t]{0.49\linewidth}
    \centering
    \includegraphics[width=\linewidth]{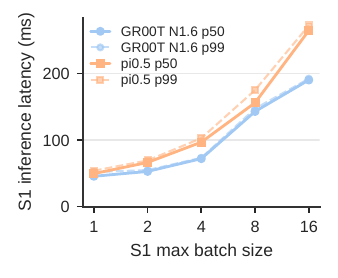}
  \end{subfigure}
  \hfill
  \begin{subfigure}[t]{0.49\linewidth}
    \centering
    \includegraphics[width=\linewidth]{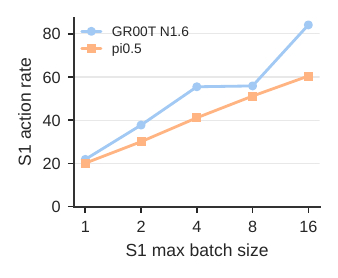}
  \end{subfigure}
    \vspace{-1em}
  \caption{System~1 performance under different batch sizes.}
    \vspace{-1em}
  \label{fig:s1_latency_throughput}
\end{figure}

\begin{figure}[t]
  \centering
  \begin{subfigure}[t]{0.49\linewidth}
    \centering
    \includegraphics[width=\linewidth]{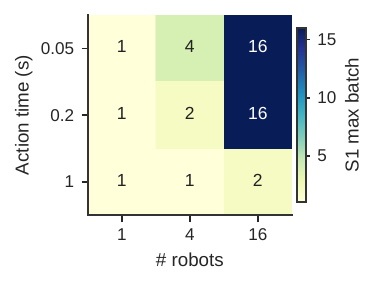}
  \end{subfigure}
  \hfill
  \begin{subfigure}[t]{0.49\linewidth}
    \centering
    \includegraphics[width=\linewidth]{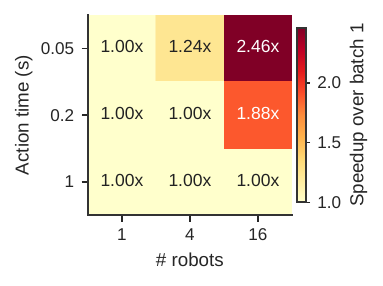}
  \end{subfigure}
    \vspace{-1em}
\caption{System~1 batching decision: the optimal batch sizes in various configurations and their speedup over batch size one.}
    \vspace{-1em}
  \label{fig:s1_batching_ablation}
\end{figure}

\begin{figure*}[t]
  \centering
  \includegraphics[width=\linewidth]{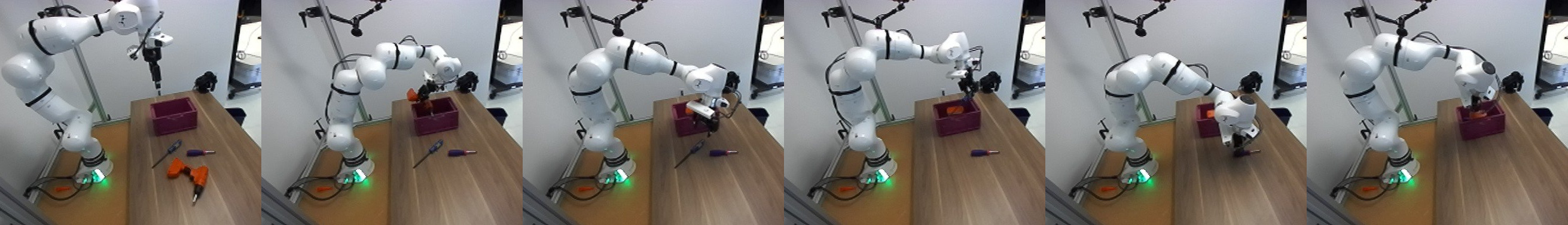}
  \caption{Real-robot execution trace: the action model drives a Franka Panda arm to place tools into a bucket.}
  \label{fig:real_robot:action}
\end{figure*}

\begin{figure*}[t]
  \centering
  \begin{subfigure}[t]{0.49\linewidth}
    \centering
    \includegraphics[width=\linewidth]{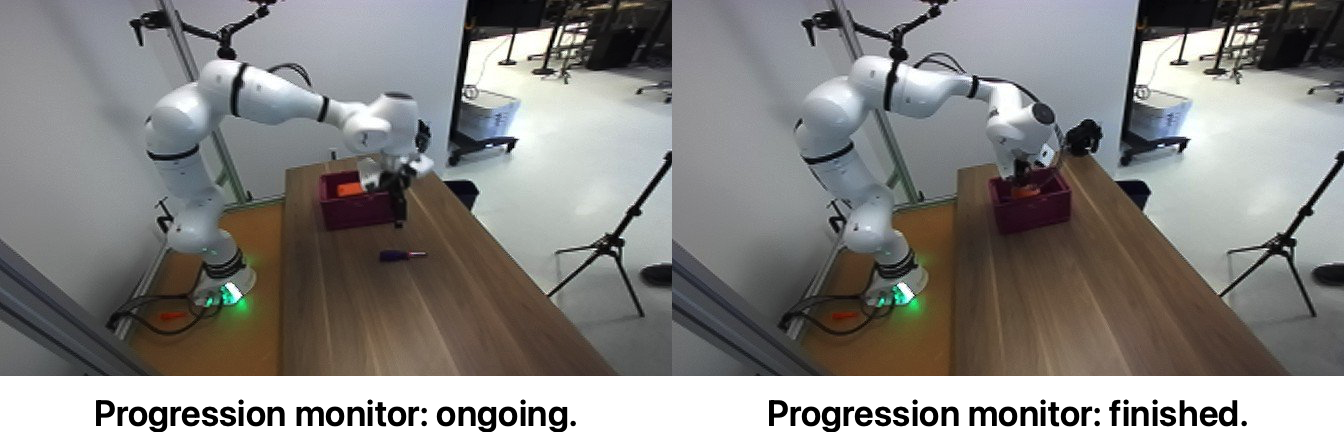}
  \end{subfigure}
  \hfill
  \begin{subfigure}[t]{0.49\linewidth}
    \centering
    \includegraphics[width=\linewidth]{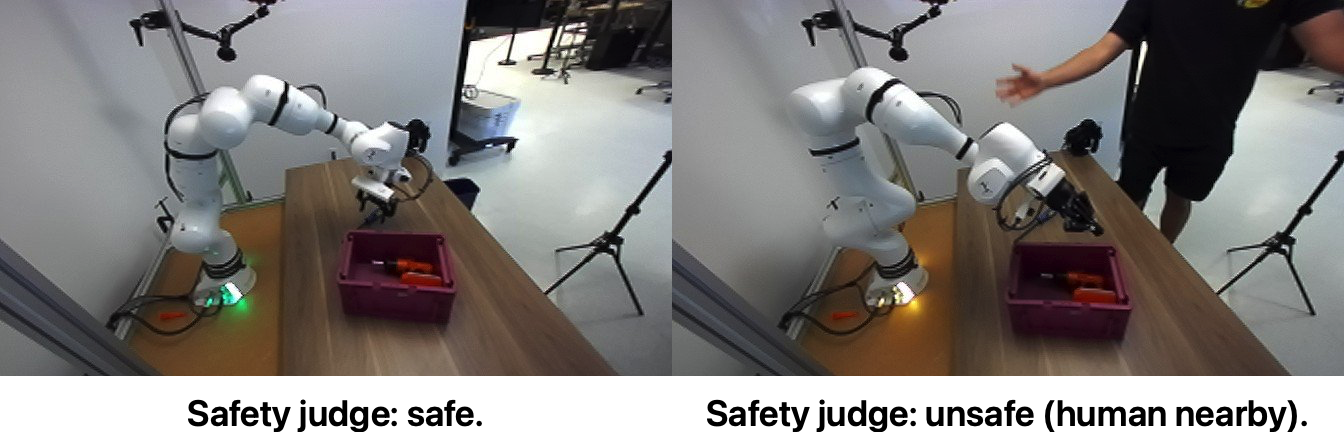}
  \end{subfigure}
  \caption{VLM judges for monitoring task completion (left) and detecting nearby humans for safe operation (right).}
  \label{fig:real_robot:vlm_judgments}
\end{figure*}

\niparagraph{Resource allocation.}
We next show that resource allocation for multi-model pipelines matters even when batching and request-rate control are carefully scheduled.
For this ablation, all schedules use the same request-rate control and System~1 batching policy derived by \ours; only the allocation of GPU servers across model components changes.
\Cref{tbl:resource_allocation_ablation} shows the resulting allocation for the P4 workload on 8 GPU servers, and \Cref{fig:allocation_ablation} shows the measured qualified throughput.
For 8 robots, \ours and the equal partitioning baseline perform similarly as there are not enough System 1 requests to saturate the GPUs yet.
However, as the number of robots increases to 16 and 32, \ours achieved 1.26$\times$ and 2.20$\times$ speedup in qualified throughput over the best static allocation, respectively. 
At 64 robots, the weighted partitioning baseline even becomes infeasible due to the insufficient capacity to meet the safety model SLO.

\niparagraph{System~1 batching.}
Finally, we isolate the effect of batching for System~1 action models.
Unlike the autoregressive VLM components, which use continuous batching, System~1 action models use a discrete batch size selected by the scheduler.
\Cref{fig:s1_latency_throughput} shows the measured latency-throughput tradeoff.
For both GR00T and \(\pi_0.5\), increasing the batch size to 16 improves inference throughput by roughly 3$\sim$4$\times$ over batch size one, at the cost of higher inference latency and additional queueing delay.

\Cref{fig:s1_batching_ablation} shows the resulting optimal batch size across robot counts and robot action periods without latency SLO: larger batches are beneficial only when the throughput gain outweighs the added delay in the robot control loop.
Under high request pressure, such as 16 robots with a 50 ms action period, the scheduler selects batch size 16 and achieves a 2.46$\times$ speedup over batch size one.
Under lighter load, such as longer action periods or fewer robots, batch size one often remains optimal because larger batches add latency without enough amortization benefit.
This motivates \ours's profiling-guided batch-size selection rather than a fixed batching policy.

\subsection{Real Robot Experiments}
\label{sec:evaluation:real_robot}

Finally, we validate that \ours can run the full robot-serving path on a Franka Panda robot arm.
Here, we ask the robot to perform a hard pick and place task similar to P3 in the synthetic workloads, where it needs to place all the tools on the table into a bucket.
\Cref{fig:real_robot:action} shows an execution trace in which the action model drives the arm to place tools into a bucket.
\Cref{fig:real_robot:vlm_judgments} then exercises the auxiliary VLM components. The task-progression monitor judges whether the task has completed or is still ongoing, while the safety model identifies whether a human has entered the unsafe region near the robot.

\section{Conclusion}
\label{sec:conclusion}

Robotics foundation models are making general-purpose robots increasingly plausible for industrial deployments.
We introduce \ours, a model-agnostic RFM serving system built around three design principles: (1) shared server-scale infrastructure for higher inference performance and efficiency, (2) robotics-aware programming abstractions and system design for multi-model orchestration, and (3) factory-objective-driven scheduling to optimize factory-level productivity.
Our evaluation shows that shared GPU-pool serving improves robot-factory productivity over shared and dedicated serving baselines by 2.44$\times$ and 12.06$\times$, respectively, while satisfying latency SLOs.
While the field is evolving rapidly, we believe these design principles are general enough to accommodate future model changes, thus providing a solid foundation for future RFM serving system designs.

\bibliographystyle{plain}
\bibliography{refs}

\end{document}